\documentclass{article}

\usepackage[dblblindworkshop, final]{neurips_2025}
\workshoptitle{Mechanistic Interpretability}

\usepackage[english]{babel}

\usepackage{amsmath}
\usepackage{graphicx}
\usepackage{hyperref}
\usepackage{makecell}

\usepackage[utf8]{inputenc} 
\usepackage[T1]{fontenc}    
\usepackage{hyperref}       
\usepackage{url}            
\usepackage{booktabs}       
\usepackage{amsfonts}       
\usepackage{nicefrac}       
\usepackage{microtype}      
\usepackage{xcolor}         

\usepackage{microtype}
\usepackage{graphicx}
\usepackage{subfigure}
\usepackage{booktabs} 
\usepackage{hyperref}

\usepackage{algorithm}

\usepackage{amsmath}
\usepackage{amssymb}
\usepackage{mathtools}
\usepackage{amsthm}

\usepackage{natbib}
\usepackage{amsmath}
\usepackage{cases}
\usepackage{graphicx}
\usepackage{tikz-cd}
\tikzcdset{scale cd/.style={every label/.append style={scale=#1},
    cells={nodes={scale=#1}}}}
\usepackage{subcaption}
\usepackage{extpfeil}
\usepackage{algorithmic}
\usepackage{comment}
\usepackage{pgfplots} 
\usepackage{enumitem}
\usepackage{svg}
\usepackage{CJKutf8}
\usepackage{adjustbox}
\usepackage{mathtools}
\usepackage{multirow}
\usepackage{listings}
\usepackage[most]{tcolorbox}
\tcbuselibrary{skins}

\definecolor{codegreen}{rgb}{0,0.6,0}
\definecolor{codegray}{rgb}{0.5,0.5,0.5}
\definecolor{codepurple}{rgb}{0.58,0,0.82}
\definecolor{backcolour}{rgb}{0.95,0.95,0.92}

\definecolor{codewhite}{rgb}{1,1,1}
\definecolor{codeblack}{rgb}{0,0,0}

\definecolor{codewhite}{rgb}{1,1,1}
\definecolor{codeblack}{rgb}{0,0,0}

\lstdefinestyle{result}{backgroundcolor=\color{codewhite},  
  commentstyle=\color{codeblack},
  keywordstyle=\color{codeblack},
  numberstyle=\tiny\color{codeblack},
  stringstyle=\color{codeblack},
  basicstyle=\ttfamily\footnotesize,
  breakatwhitespace=false,     
  breaklines=true,             
  captionpos=b,                
  keepspaces=true,             
  numbers=left,                
  numbersep=5pt,              
  showspaces=false,           
  showstringspaces=false,
  showtabs=false,             
  tabsize=2
}
\definecolor{myred}{rgb}{1, 0, 0}    
\definecolor{mygreen}{rgb}{0, 0.5, 0} 
\definecolor{myblue}{rgb}{0, 0, 1}   

\definecolor{c1}{rgb}{1, 0, 0}    
\definecolor{c2}{rgb}{0, 0.8, 0}   
\definecolor{c3}{rgb}{0, 0, 1}   
\definecolor{c4}{rgb}{0.5, 0, 0.5}   

\definecolor{c11}{rgb}{1, 0, 0}    
\definecolor{c12}{rgb}{0.8, 0.5, 0}   
\definecolor{c13}{rgb}{0, 0.8, 0}   
\definecolor{c14}{rgb}{0, 0.5, 0.5}   
\definecolor{c15}{rgb}{0, 0.2, 1}   
\definecolor{c16}{rgb}{0.5, 0, 0.5}   

\theoremstyle{plain}
\usepackage[bb=ams]{mathalpha}

\theoremstyle{definition}

\theoremstyle{remark}

\title{Unveiling the Latent Directions of Reflection \\ in Large Language Models}
\author{Fu-Chieh Chang$^{1,2}$\quad Yu-Ting Lee$^{2}$\quad Pei-Yuan Wu$^{2,3}$ \\
$^{1}$MediaTek Research, Taipei, Taiwan \\
$^{2}$Graduate Institute of Communication Engineering, National Taiwan University, Taipei, Taiwan  \\
 $^{3}$AI Research Center (AINTU), National Taiwan University, Taipei, Taiwan \\
\texttt{d09942015@ntu.edu.tw}
,~\texttt{r14942088@g.ntu.edu.tw}
,~\texttt{peiyuanwu@ntu.edu.tw}
}
\date{}

\begin{document}
\maketitle
\begin{abstract}
Reflection, the ability of large language models (LLMs) to evaluate and revise their own reasoning, has been widely used to improve performance on complex reasoning tasks. Yet,  most prior works emphasizes designing reflective prompting strategies or reinforcement learning objectives, leaving the inner mechanisms of reflection underexplored. In this paper, we investigate reflection through the lens of latent directions in model activations. We propose a methodology based on activation steering to characterize how instructions with different reflective intentions: no reflection, intrinsic reflection, and triggered reflection. By constructing steering vectors between these reflection levels, we demonstrate that (1) new reflection-inducing instructions can be systematically identified, (2) reflective behavior can be directly enhanced or suppressed through activation interventions, and (3) suppressing reflection is considerably easier than stimulating it. Experiments on GSM8k-adv and Cruxeval-o-adv with Qwen2.5-3B and Gemma3-4B-IT reveal clear stratification across reflection levels, and steering interventions confirm the controllability of reflection. Our findings highlight both opportunities (e.g., reflection-enhancing defenses) and risks (e.g., adversarial inhibition of reflection in jailbreak attacks). This work opens a path toward mechanistic understanding of reflective reasoning in LLMs.
\end{abstract}

\section{Introduction}
Large language models (LLMs) have demonstrated remarkable capabilities in multi-step reasoning tasks~\cite{wei2022chain, kojima2022large, huang-chang-2023-towards, plaat2024reasoning}, with reflection playing a central role in their success~\cite{ai2025rethinkingreflectionpretraining,bo2024reflective,yuan2025reinforcellmreasoningmultiagent,shinn2023reflexionlanguageagentsverbal}. Reflection enables a model to reconsider its reasoning process, identify errors, and revise its conclusions, thereby producing more reliable outputs~\cite{ai2025rethinkingreflectionpretraining,shinn2023reflexionlanguageagentsverbal}. While reflection has been operationalized in diverse ways, such as multi-agent frameworks~\cite{bo2024reflective,yuan2025reinforcellmreasoningmultiagent}, long chain-of-thought prompting~\cite{chen2025reasoningerasurveylong,wang2025r1compresslongchainofthoughtcompression,ji2025mygomultiplexcotmethod}, and iterative refinement~\cite{shinn2023reflexionlanguageagentsverbal,madaan2023selfrefineiterativerefinementselffeedback,pan-etal-2024-automatically}, the mechanisms underlying how reflection emerges in LLMs remain poorly understood. Most prior research has treated reflection as a behavioral property to be exploited, rather than as a latent phenomenon to be explained. A detailed literature study is presented in Appendix~\ref{sec:related_works}.
In this paper, we move beyond behavioral prompting strategies and instead focus on the mechanistic interpretability of reflection. Building on recent advances in activation steering~\cite{turner2024steeringlanguagemodelsactivation,zou2025representationengineeringtopdownapproach,2024_Refusal}, we investigate whether reflection aligns latent directions in a model’s hidden space. Our contributions are summarized as follows:
\begin{itemize}
    \item We categorize reflection into three levels: No Reflection, Intrinsic Reflection, and Triggered Reflection. This stratification enables the construction of steering vectors that capture the latent transitions between different reflective states.
    \item Using these steering vectors, we demonstrate a principled approach to discovering new reflection-inducing prompts, moving beyond trial-and-error prompt design.
    \item We show that reflective behavior can be directly modulated through activation steering, enabling both enhancement and inhibition of reflection at inference time.
    \item Our findings reveal an asymmetry: suppressing reflection is easier than inducing it. This observation not only sheds light on the underlying mechanisms of reasoning but also raises potential security concerns, as malicious actors could exploit reflection inhibition to bypass model safeguards.
\end{itemize}

\begin{table}[h]
    \centering
        \caption{List of Instruction for Reflection in Related Works}
\begin{small}
\begin{tabular}{|l|l|}
\hline 
Words & Source \\
\hline 
    wait & \cite{ai2025rethinkingreflectionpretraining} \\ \hline 
    wait & \cite{muennighoff2025s1simpletesttimescaling} \\ \hline
    wait, alternatively, double-check, 
make sure, another way, verify, to
confirm & \cite{zhang2025reasoningmodelsknowtheyre}\\
    \hline 
    wait, alternatively & \cite{chen2025sealsteerablereasoningcalibration} \\
    \hline
    wait, alternatively, recheck, retry, however & \cite{yeo2025demystifyinglongchainofthoughtreasoning} \\ \hline
     wait, alternatively, double-check, let me check, emm, hmm &  \cite{wang2025r1compresslongchainofthoughtcompression} \\ \hline
\end{tabular}
\end{small}
    \label{tab:trigger}
\end{table}

\section{Methodology}
\subsection{Problem Formulation}
We begin by defining the type of \textbf{Reflection} considered in this work. Specifically, we focus on \textbf{Situational Reflection}~\cite{ai2025rethinkingreflectionpretraining}, where a model reflects on reasoning generated by another source (e.g., a different model). Other forms, such as \textbf{Self-reflection}—where a model critiques its own outputs—are beyond our scope. We choose situational reflection because it provides a more controlled setting to study how models correct deliberately induced errors. Importantly, in our formulation the errors are introduced within the \emph{reasoning steps} rather than only in the final answer. This contrasts with works like~\cite{lee2025explanationintrinsicselfcorrectionlinear}, which focus on correcting end outputs in non-reasoning tasks.
An illustrative example, adapted from~\cite{ai2025rethinkingreflectionpretraining}, is shown in Figure~\ref{fig:level_of_reflection}: the prompt presents a GSM8k math problem, followed by a deliberately flawed chain-of-thought, and ends with an instruction to trigger reflection (e.g., \emph{wait}, \emph{alternatively}). 
Prior work has proposed various trigger instructions (summarized in Table~\ref{tab:trigger}), but their selection has largely  based on intuition rather than systematic analysis. This raises two key research questions:  
\begin{itemize}
    \item How can we systematically identify effective trigger instructions, rather than relying on trial-and-error?  
    \item Do effective trigger instructions correspond to latent directions in the hidden space that implicitly induce the self-reflection process?  
\end{itemize}

To address these questions, we propose a methodology grounded in \emph{activation steering}. Specifically, we first categorize reflection into three levels—\emph{No Reflection}, \emph{Intrinsic Reflection}, and \emph{Triggered Reflection}—to establish a structured framework for analysis. We then compute \emph{steering vectors} between these levels, capturing the latent directions that separate different reflective behaviors. These steering vectors serve two purposes: \begin{itemize}
    \item They allow us to discover new instructions beyond those reported in prior work by comparing the alignment of candidate tokens with the reflection-related steering direction.
    \item They enable controlled interventions, where reflection can be enhanced or inhibited by adding or subtracting steering vectors at selected layers.
\end{itemize}

\begin{figure}[h]
    \centering    
\begin{small}

\begin{tcolorbox}[breakable, enhanced jigsaw, opacityback=0,boxsep=1.5pt,left=1.5pt,right=1.5pt,top=1.5pt,bottom=1.5pt]
\textbf{Prompt:} 
Answer the question:
John drives for 3 hours at a speed of 60 mph and then turns around 
because he realizes he forgot something very important at home. 
He tries to get home in 4 hours but spends the first 2 hours in standstill traffic. 
He spends the next half-hour driving at a speed of 30mph, 
before being able to drive the remaining time of the 4 hours going at 80 mph. 
How far is he from home at the end of those 4 hours?
Please always end your response with the final numerical answer

\vspace{1.5mm}

Let’s solve this step by step … 

\vspace{1.5mm}

When he turned around he was 3*60=<<3*60=180>>180 miles from home,
He was only able to drive 4-2=<<4-2=2>>2 hours in the first four hours.
In half an hour he goes 30*.5=<<30*.5=15>>15 miles.
He then drives another 2-.5=<<2-.5=1.5>>1.5 hours.
In that time he goes 80*1.5=<<80*1.5=120>>120 miles.
So he drove 120+15=<<120+15=135>>135 miles.
So he is still 180=<<180=180>>180 miles away from home. \textbf{[Instruction]}

\vspace{2.5mm}

\textbf{Ground-Truth:} 45

\vspace{2.5mm}

\begin{tabular}{|l|l|l|l|}
\hline
\textbf{Level of Reflection} & \textbf{Instruction} & \textbf{Response} & \textbf{Correctness} \\
\hline No Reflection
& Answer
& 180
& False \\
\hline Intrinsic Reflection 
& [EOS]
& 
Calculate the distance ... , 120 miles from home.
& False \\
\hline Triggered Reflection
& Wait
& 180-135=<<180-135=45>>45 miles
& True
\\
\hline
\end{tabular}
\end{tcolorbox}
\end{small}
\caption{An example of reflection, adapted from~\cite{ai2025rethinkingreflectionpretraining}.}
\label{fig:level_of_reflection}
\end{figure}

\subsection{Three Levels Reflection}
In this section, we formalize the different levels of reflection.  Unlike prior works (e.g.,~\cite{ai2025rethinkingreflectionpretraining, zhang2025reasoningmodelsknowtheyre}) which explicitly give instructions to trigger the act of reflection, we emphasize that reflection can occur even without explicit triggers: LLMs sometimes spontaneously revise their reasoning, a behavior we call \emph{intrinsic reflection}.  This usually perform worse than explicitly triggered reflection. Conversely, reflection can also be suppressed entirely by instructing the model to output an answer directly after a flawed chain-of-thought, a case we call \emph{no reflection}.  
In summary, we distinguish three levels of reflection: \emph{No Reflection}, \emph{Intrinsic Reflection}, and \emph{Triggered Reflection}. Figure~\ref{fig:level_of_reflection} illustrates these cases, where different instructions lead to distinct behaviors:
\begin{itemize}
    \item \textbf{No Reflection}: when the model is forced to answer immediately (e.g., \emph{Answer}), it simply outputs the conclusion from the flawed reasoning without revision.
    \item \textbf{Intrinsic Reflection}: when the instruction has no intention to trigger or stop reflection (e.g., \emph{[EOS]}), the model continues its chain-of-thought, which may or may not correct earlier errors.
    \item \textbf{Triggered Reflection}: when given an explicit cue (e.g., \emph{Wait}), the model inspects its reasoning steps and often revises them to produce the correct answer.
\end{itemize}

This stratification allows us to study different levels of reflection induced by prompts with different intentions. It also enables us to define the steering vector as a contrastive latent direction that encodes the difference between two reflection behaviors in activation space, for example, the vector from “No Reflection” to “Triggered Reflection.” In practice, this steering vector is computed as the average activation difference at a specific layer between samples exhibiting the respective behaviors.

\subsection{Latent Directions of Reflection}
Having defined the three levels of reflection, we use contrastive pairs to construct steering vectors between them. We also note that extracting steering vectors from contrastive pairs is an established method~\cite{turner2024steeringlanguagemodelsactivation,zou2025representationengineeringtopdownapproach,rimsky-etal-2024-steering}. Let $I_{2}$, $I_{1}$, and $I_{0}$ denote the sets of instructions corresponding to \emph{Triggered Reflection}, \emph{Intrinsic Reflection}, and \emph{No Reflection}, respectively. We consider a dataset $\mathcal{D}$ of reasoning problems with deliberately flawed chain-of-thoughts. We first sample a subset of training data from $\mathcal{D}$, denoted as $\mathcal{D}_{train}$. We then select a pair of levels $(a,b)$ with $b>a$, where $I_a$ and $I_b$ are the corresponding instruction sets. For each sample $d \in \mathcal{D}_{train}$, we append instructions $i_b \in I_b$ and $i_a \in I_a$ to form augmented prompts $d_{i_b}$ and $d_{i_a}$. Here, $b$ corresponds to the reflection-inducing instruction set, while $a$ serves as the reference baseline.  
For instance, in Fig.~\ref{fig:level_of_reflection}, the sample $d$ consists of a GSM8k problem accompanied by a chain-of-thought containing deliberate errors. A portion of a sample $d$ is shown below.

\begin{small}
\begin{tcolorbox}[breakable, enhanced jigsaw, opacityback=0,boxsep=1.5pt,left=1.5pt,right=1.5pt,top=1.5pt,bottom=1.5pt]
\textbf{Prompt:} Answer the question: John drives for 3 hours at a speed of 60 mph and then turns
around because he realizes he forgot something very important at home ....  
Let’s solve this step by step ...  
2-.5=«2-.5=1.5»1.5 hours. In that time he  
goes 80*1.5=«80*1.5=120»120 miles. So he drove 120+15=«120+15=135»135 miles. So he  
is still 180=«180=180»180 miles away from home. 
\end{tcolorbox}
\end{small}
After appending the instruction $i_2=\text{`Wait'}$ from $I_2$, the modified prompt $d_{i_2}$ becomes:  

\begin{small}
\begin{tcolorbox}[breakable, enhanced jigsaw, opacityback=0,boxsep=1.5pt,left=1.5pt,right=1.5pt,top=1.5pt,bottom=1.5pt]
\textbf{Prompt:} Answer the question: John drives for 3 hours at a speed of 60 mph and then turns
around because he realizes he forgot something very important at home ....  
Let’s solve this step by step ...  
2-.5=«2-.5=1.5»1.5 hours. In that time he  
goes 80*1.5=«80*1.5=120»120 miles. So he drove 120+15=«120+15=135»135 miles. So he  
is still 180=«180=180»180 miles away from home. \textbf{Wait}
\end{tcolorbox}
\end{small}
This appended instruction provides a controlled signal to the model that determines whether it explicitly reflects, halts, or ignore the instruction and then continues reasoning. By contrasting activations from different reflection levels, we can extract a \emph{latent direction of reflection} in the hidden space.  For a given LLM and layer $\ell \in [L]$, we compute the mean activation at the token position of the appended instruction:
$$
\boldsymbol{\mu}_{i_b}^{(\ell)}=\frac{1}{|\mathcal{D}_{train}|} \sum_{d \in \mathcal{D}_{train}} \mathbf{x}^{(\ell)}(d_{i_b}), 
\quad
\boldsymbol{\mu}_{i_a}^{(\ell)}=\frac{1}{|\mathcal{D}_{train}|} \sum_{d \in \mathcal{D}_{train}} \mathbf{x}^{(\ell)}(d_{i_a}).
$$  
where $\mathbf{x}^{(\ell)}(d)$ is the activation of the LLM at layer $\ell$ given input $d$.  In our work, $\mathbf{x}^{(\ell)}(d)$  is defined as the output of the $\ell$-th attention block, computed by adding the MLP output (following the $\ell$-th attention layer) to the residual stream.
The steering vector from level $a$ to level $b$ at layer $\ell$ is then defined as:
$$
\boldsymbol{\mu}_{a\rightarrow b}^{(\ell)}=
\frac{1}{|I_a||I_b|}\sum_{i_b\in I_b}\sum_{i_a\in I_a}
\big(\boldsymbol{\mu}^{(\ell)}_{i_b}-\boldsymbol{\mu}^{(\ell)}_{i_a}\big).
$$  

Intuitively, $\boldsymbol{\mu}_{a\rightarrow b}^{(\ell)}$ captures the latent shift in hidden representations required to move the model’s behavior from level $a$ (e.g., \emph{No Reflection}) toward level $b$ (e.g., \emph{Triggered Reflection}). These vectors provide a principled way to both discover new trigger instructions and intervene in the model’s reflective behavior.

\subsubsection{Steering Vectors for Discovering New Instructions}

We define a candidate pool of instructions $I'$ that are not included in the original sets $I_0$, $I_1$, or $I_2$, but may potentially serve as reflection triggers as in $I_2$.  
The key idea is to test whether these new instructions exhibit activation patterns aligned with known reflection-inducing instructions.  
To do so, we compare the steering vector induced by each $i' \in I'$ against the canonical steering direction $\mu^{(\ell)}_{a\rightarrow b}$ derived from established reflection levels.  
To evaluate whether a new instruction $i' \in I'$ behaves similarly to instructions in $I_b$, we compute its steering vector relative to $I_a$:
$$
\boldsymbol{\mu}_{a\rightarrow i'}^{(\ell)}=
\frac{1}{|I_a|}\sum_{i_a\in I_a}
\big(\boldsymbol{\mu}^{(\ell)}_{i'}-\boldsymbol{\mu}^{(\ell)}_{i_a}\big).
$$  
We then measure the cosine similarity between $\boldsymbol{\mu}_{a\rightarrow i'}^{(\ell)}$ and $\boldsymbol{\mu}_{a\rightarrow b}^{(\ell)}$, denoted as  
$$
\operatorname{CosSim}(\mu^{(\ell)}_{a\rightarrow i'},\mu^{(\ell)}_{a\rightarrow b})
=\frac{\left(\mu^{(\ell)}_{a\rightarrow i'}\right)^{\top}\mu^{(\ell)}_{a\rightarrow b}}
{\|\mu^{(\ell)}_{a\rightarrow i'}\|_2\|\mu^{(\ell)}_{a\rightarrow b}\|_2}.
$$

A similarity value close to $1$ indicates that the candidate instruction $i'$ activates the model’s hidden space in a manner consistent with the reflection-inducing instructions in $I_b$, and thus has potential to serve as a new reflection trigger.  
The choice of reference level $a$ and target level $b$, as well as the layer $\ell$ at which similarity is computed, are determined empirically and discussed in Sec.~\ref{sec:steer_new_inst}.

\subsubsection{Steering Vectors for Intervening in Reflection}
Beyond discovering new instructions, steering vectors can also be used to directly \emph{control} the reflective behavior of LLMs. For a given layer $\ell \in [L]$, let $\mathbf{x}^{(\ell)}(d)$ denote the activation at layer $\ell$ for input $d$. We consider two complementary modes of intervention:  

\begin{itemize}
    \item \textbf{Enhancing Reflection:}  
    To strengthen reflection, we apply the steering vector in the forward direction (from a lower level $a$ to a higher level $b$):  
    $
    \mathbf{x}^{(\ell)}(d) \;\leftarrow\; \mathbf{x}^{(\ell)}(d) + \boldsymbol{\mu}_{a\rightarrow b}^{(\ell)}.
    $
    \item \textbf{Inhibiting Reflection:}  
    To suppress reflection, we apply the reverse direction. Noting that $-\boldsymbol{\mu}_{b\rightarrow a}^{(\ell)} = \boldsymbol{\mu}_{a\rightarrow b}^{(\ell)}$, we intervene as:  
    $
    \mathbf{x}^{(\ell)}(d) \;\leftarrow\; \mathbf{x}^{(\ell)}(d) + \boldsymbol{\mu}_{b\rightarrow a}^{(\ell)}.
    $
\end{itemize}

During inference, the intervention is applied only once—at a single layer $\ell$, and specifically at the token position corresponding to the appended instruction in $d$.  
The optimal choice of intervention layer $\ell$ is determined empirically, as discussed in Sec.~\ref{sec:steer_intervene}.

\section{Experiment and Result}
\subsection{Experiment Setup}
We conduct experiments to validate our proposed approach. For the models, we select Qwen2.5-3B~\cite{qwen2025qwen2.5} and Gemma3-4B-IT~\cite{gemma32025}, as they strike a balance between computational tractability and reasoning performance. These LLMs are large enough to exhibit reflective behaviors while still lightweight enough to allow systematic interventions and multiple runs across datasets.
For evaluation, we use the dataset \texttt{gsm8k\_adv} and \texttt{cruxeval\_o\_adv} introduced in~\cite{ai2025rethinkingreflectionpretraining}. Accuracy is computed as the proportion of samples whose predicted answers exactly match the ground-truth. To ensure robustness, we apply a flexible extraction procedure: if the model’s response contains a number exactly matching the ground-truth, it is counted as correct. Further experimental details are provided in a  repository\footnote{\url{https://github.com/d09942015ntu/unveiling_directions_reflection}}.

\begin{table}[h]
\caption{Results across three levels of reflection. Each entry reports exact-match accuracy under different reflection-inducing instructions or the average accuracy within a given level.}
\centering
\resizebox{\textwidth}{!}{
\begin{scriptsize}
\begin{tabular}{|c|c|c|c|c|c|}
\hline
\multicolumn{1}{|l|}{} & Instruction   & \begin{tabular}[c]{@{}c@{}}Qwen2.5-3B \\ \texttt{gsm8k\_adv} \end{tabular}  & \begin{tabular}[c]{@{}c@{}}Gemma3-4B-IT \\ \texttt{gsm8k\_adv} \end{tabular}  & \begin{tabular}[c]{@{}c@{}}Qwen2.5-3B \\ \texttt{cruxeval\_o\_adv} \end{tabular} & \begin{tabular}[c]{@{}c@{}}Gemma3-4B-IT \\ \texttt{cruxeval\_o\_adv} \end{tabular}  \\ \hline\hline
\multirow{4}{*}{Triggered Reflection} 
& \emph{Wait}          & .360           & .587  & .090           & .212             \\ \cline{2-6} 
& \emph{Alternatively} & .470           & .684 & .076           & .270               \\ \cline{2-6} 
& \emph{Check}         & .363           & .537 & .030           & .258              \\ \cline{2-6} 
& Average              & .397           & .586 & .065           & .247              \\ \hline\hline
\multirow{4}{*}{Intrinsic Reflection} 
& \emph{[EOS]}         & .328           & .252 & .046           & .038              \\ \cline{2-6} 
& \emph{\#}            & .281           & .327 & .040           & .216              \\ \cline{2-6} 
& \emph{\%}            & .278           & .428 & .034           & .114              \\ \cline{2-6} 
& Average              & .295           & .335  & .040           & .133               \\ \hline\hline
\multirow{4}{*}{No Reflection}        
& \emph{Answer}        & .037           & .157 & .002           & .052               \\ \cline{2-6} 
& \emph{Result}        & .071           & .206 & .028           & .058               \\ \cline{2-6} 
& \emph{Output}        & .046           & .079 & .022           & .044               \\ \cline{2-6}
& Average              & .051           & .147 & .017           & .051               \\ \hline
\end{tabular}
\label{table:three_level_result}
\end{scriptsize}
}
\end{table}

\subsection{Three Levels of Reflection}
In this experiment, we examine how LLMs respond to instructions designed with three distinct intentions:  
(1) \emph{Trigger Reflection}, where explicit cues encourage the model to revisit and refine its reasoning;  
(2) \emph{Intrinsic Reflection}, where semantically neutral tokens provide no explicit guidance but still allow spontaneous continuation of reasoning; and  
(3) \emph{No Reflection}, where direct-answer instructions suppress further reasoning and force immediate output.  
To trigger reflection, we select three of the most common reflective cues—\emph{wait}, \emph{alternative}, and \emph{check}—from Table~\ref{tab:trigger}. 
For simplicity and without loss of generality, multi-token variants such as \emph{double-check} and \emph{recheck} are treated as the single token \emph{check}.
For intrinsic reflection, we employ instructions without inherent semantic intent, such as the \emph{[EOS]} token or symbols like \emph{"\%"} and \emph{"\#"}.
Finally, to enforce no reflection, we adopt direct-answer instructions—\emph{Answer}, \emph{Output}, and \emph{Result}—which explicitly request final responses without revisiting prior reasoning.  

\paragraph{Result:}
Table~\ref{table:three_level_result} reports the accuracy under each condition. On average, triggered reflection yields the best performance, followed by intrinsic reflection, and no reflection performs worst. This clear stratification validates our hypothesis that the LLMs has three different levels of reflection behavior when prompting with instruction with different type of intentions. This result also demonstrates that explicit reflective cues substantially improve reasoning reliability compared to neutral or suppressive instructions. Illustrative examples of prompts and corresponding responses under each instruction for \texttt{gsm8k\_adv} dataset are presented in Sec.~\ref{sec:exp_sample_question_response}.

\subsection{Steering Vector to Discover New Instructions}
\label{sec:steer_new_inst}

We build upon the three instruction sets introduced earlier: 
$I_2=\{\text{Wait}, \text{Alternatively}, \text{Check}\}$, 
$I_1=\{\text{[EOS]}, \#, \%\}$, and 
$I_0=\{\text{Answer}, \text{Result}, \text{Output}\}$.  
For our analysis, we fix $b=2$ and $a\in \{0,1\}$, and compute steering vectors $\boldsymbol{\mu}_{0\rightarrow2}$ and $\boldsymbol{\mu}_{1\rightarrow2}$ using the training subset $\mathcal{D}_{train}$ from \texttt{gsm8k\_adv}. To discover novel reflection-inducing instructions, we define a candidate pool $I'$ consisting of English vocabulary tokens drawn from the Qwen2.5 and Gemma3 tokenizer. We normalize these candidates using stemming and lemmatization (via the NLTK package~\cite{bird2009natural}). For each candidate instruction $i' \in I'$, we compute its steering vector relative to $I_a$ and measure its cosine similarity with the ground-truth steering vector, namely   
$
\operatorname{CosSim}(\boldsymbol{\mu}^{(\ell)}_{a\rightarrow i'}, \boldsymbol{\mu}^{(\ell)}_{a\rightarrow 2})$ where $a\in\{0,1\}$.
Candidates with the highest similarity are hypothesized to function as new reflection triggers. We then rank instructions by similarity and select the top-3, top-5 and top-8 candidates for evaluation on a held-out $\mathcal{D}_{test}$ split. Their effectiveness is assessed by appending the candidate instruction to each problem and measuring accuracy on \texttt{gsm8k\_adv} and \texttt{cruxeval\_o\_adv}. As a baseline, we also compare against candidate selection based purely on input embedding of cosine similarity of instruction tokens from $I_2$, without using steering vectors.  

\begin{figure}
    \centering
    \begin{tabular}{c c c c}
    \input{fig/exp3_MyQwen2.5-3B_gsm8k_adv_3}
    & \input{fig/exp3_gemma-3-4b-it_gsm8k_adv_3}
    & \input{fig/exp3_MyQwen2.5-3B_cruxeval_o_adv_3}
    & \input{fig/exp3_gemma-3-4b-it_cruxeval_o_adv_3}
    \\
    \input{fig/exp3_MyQwen2.5-3B_gsm8k_adv_5}
    & \input{fig/exp3_gemma-3-4b-it_gsm8k_adv_5}
    & \input{fig/exp3_MyQwen2.5-3B_cruxeval_o_adv_5}
    & \input{fig/exp3_gemma-3-4b-it_cruxeval_o_adv_5}
    \\
    \input{fig/exp3_MyQwen2.5-3B_gsm8k_adv_8}
    & \input{fig/exp3_gemma-3-4b-it_gsm8k_adv_8}
    & \input{fig/exp3_MyQwen2.5-3B_cruxeval_o_adv_8}
    & \input{fig/exp3_gemma-3-4b-it_cruxeval_o_adv_8}
    \\
    & & \\
    \end{tabular}
    \begin{tabular}{|c|}
    \hline
     \textcolor{c11}{---}~$\boldsymbol{\mu}_{0\rightarrow2}$ \quad 
     \textcolor{c12}{---}~$\boldsymbol{\mu}_{1\rightarrow2}$ \quad  
     \textcolor{c13}{$\boldsymbol{\cdots}$} $I_2$ \quad 
     \textcolor{c14}{$\boldsymbol{\cdots}$} $I_1$\quad 
     \textcolor{c15}{$\boldsymbol{\cdots}$} $I_0$ \quad 
     \textcolor{c16}{- -}~ Input Embedding \\
    \hline
    \end{tabular}
    \caption{Average accuracy of discovered instructions ranked by cosine similarity with steering vectors across layers $\ell$, compared against the average accuracy of instructions in $I_2$, $I_1$, and $I_0$, as well as instructions selected based on input embedding similarity to $I_2$.}
\label{fig:instruction_discovery}
\end{figure}

\begin{table}[h]
    \centering
    \caption{Top-5 example instructions with their cosine similarity (to either the steering vector or input embedding) and corresponding performance on \texttt{gsm8k\_adv}. (Left: Qwen2.5-3B, Right: Gemma3-4B-IT)}
    \label{tab:instruction_discovery}
    \resizebox{\textwidth}{!}{
    \begin{tiny}
\begin{tabular}{|c|c|c|c||c|c|c|c|} 
\hline 
Vector & Instruction & CosineSim & Accuracy & Vector & Instruction & CosineSim & Accuracy \\ 
\hline 
\multirow{6}{*}{{\begin{tabular}[c]{@{}c@{}}Input\\ Embed\end{tabular}} } & Additionally & 0.6600 & 0.4415 &\multirow{6}{*}{{\begin{tabular}[c]{@{}c@{}}Input\\ Embed\end{tabular}} } & Verify & 0.5586 & 0.5315 \\ 
 & Verify & 0.6390 & 0.3765 & & Additionally & 0.5378 & 0.5740 \\ 
 & Await & 0.6255 & 0.3765 & & Look & 0.5258 & 0.5435 \\ 
 & ConfigureAwait & 0.6230 & 0.0930 & & Furthermore & 0.5191 & 0.5880 \\ 
 & Unchecked & 0.5904 & 0.2405 & & Watch & 0.5187 & 0.3910 \\ 
 \cline{2-4}  \cline{6-8} & Averaged & 0.6276 & 0.2777 &  & Averaged & 0.5320 &  0.5256\\ 

\hline 
\multirow{6}{*}{$\boldsymbol{\mu}_{0\rightarrow 2}^{(12)}$} & Verify & 0.6291 & 0.3765 &\multirow{6}{*}{$\boldsymbol{\mu}_{0\rightarrow 2}^{(12)}$} & Verify & 0.9780 & 0.5315 \\ 
 & However & 0.6083 & 0.4500 & & Confirm & 0.9639 & 0.5205 \\ 
 & Then & 0.6060 & 0.4615 & & Initially & 0.9613 & 0.5835 \\ 
 & Otherwise & 0.6022 & 0.4445 & & Oops & 0.9577 & 0.7030 \\ 
 & Meanwhile & 0.6000 & 0.3990 & & Validate & 0.9563 & 0.4880 \\ 
  \cline{2-4}  \cline{6-8} & Averaged & 0.6091 & \textbf{0.4263} &  & Averaged & 0.9634 & \textbf{0.5653} \\ 
\hline 
\multirow{6}{*}{$\boldsymbol{\mu}_{1\rightarrow 2}^{(12)}$} & Verify & 0.6673 & 0.3765 &\multirow{6}{*}{$\boldsymbol{\mu}_{1\rightarrow 2}^{(12)}$} & Verification & 0.9919 & 0.5170 \\ 
 & Look & 0.6488 & 0.2670 & & Confirmation & 0.9909 & 0.5040 \\ 
 & Alternate & 0.6136 & 0.3970 & & Oops & 0.9883 & 0.7030 \\ 
 & Await & 0.6125 & 0.2370 & & Validation & 0.9882 & 0.5155 \\ 
 & Otherwise & 0.6097 & 0.4445 & & Initially & 0.9870 & 0.5835 \\
   \cline{2-4}  \cline{6-8} & Averaged & 0.6304 & 0.3444 &  & Averaged & 0.9893 & 0.5645 \\ 
\hline 
\end{tabular}\end{tiny}

    }
\end{table}

\paragraph{Results:} 
Figure~\ref{fig:instruction_discovery}  report averaged accuracy for top-3, top-5 and top-8 instructions, and Table~\ref{tab:instruction_discovery} shows examples of top-5 instructions for $\ell=12$. For a clear comparison, Figure~\ref{fig:instruction_discovery} also presents the average accuracy of instructions in $I_2$, $I_1$ and $I_0$, represented by dotted
line.. We make three observations:  
\begin{itemize}
    \item Steering vectors derived from $\boldsymbol{\mu}_{0\rightarrow2}$ slightly outperform those from $\boldsymbol{\mu}_{1\rightarrow2}$, suggesting that contrasting No Reflection with Triggered Reflection provides a stronger signal than contrasting Intrinsic with Triggered Reflection.  
    \item Reflection-inducing directions emerge more clearly in higher layers ($\ell > 5$), consistent with the intuition that reflective reasoning requires late-stage integration of semantic and reasoning signals.  
    \item Baselines using only input embedding similarity often select semantically related but non-reflective tokens (e.g., \emph{Await}, \emph{ConfigureAwait}, \emph{Unchecked}), which fail to improve accuracy. By contrast, steering vectors discover effective triggers such as \emph{However} and \emph{Otherwise}, which align with instructions previously reported in reflective datasets (Table~\ref{tab:trigger}).  
\end{itemize}

These results demonstrate that steering vectors capture latent directions of reflection more faithfully than surface-level embedding similarity, enabling systematic discovery of reflection-inducing instructions. 

\subsection{Steering Vectors for Intervening in Reflection}  
\label{sec:steer_intervene}  
To study how reflection can be modulated, we apply steering vectors in two complementary directions:  

\begin{itemize}
    \item \textbf{Enhancing Reflection.} We apply $\boldsymbol{\mu}_{0\rightarrow 2}$ and $\boldsymbol{\mu}_{0\rightarrow 1}$ to samples appended with the instructions \emph{[EOS]} and \emph{Answer}, respectively. These interventions are designed to push the model’s activations toward stronger reflective behavior.  

    \item \textbf{Inhibiting Reflection.} We apply $\boldsymbol{\mu}_{2\rightarrow 0}$ and $\boldsymbol{\mu}_{1\rightarrow 0}$ to samples appended with the instructions \emph{Wait} and \emph{[EOS]}, respectively. These interventions are intended to suppress reflection, encouraging the model to terminate reasoning prematurely.  
\end{itemize}

The steering vectors are computed using $\mathcal{D}_{train}$, while the effects of enhancement and inhibition are evaluated on $\mathcal{D}_{test}$. We conduct experiments on two datasets: \texttt{gsm8k\_adv} and \texttt{cruxeval\_o\_adv} introduced by~\cite{ai2025rethinkingreflectionpretraining}. Performance is reported as the percentage of questions answered correctly. 
We perform activation steering across model layers $\ell$ and report the resulting accuracy after applying the steering vector at each layer.

\paragraph{Results:}  
Figure~\ref{fig:pos_intervention} shows the results of enhancing reflection, while Figure~\ref{fig:neg_intervention} shows the results of inhibiting reflection. For clarity, we report the average accuracy of $I_2$, $I_1$, and $I_0$, along with the baseline accuracies of \emph{Wait}, \emph{[EOS]}, and \emph{Answer} without intervention. For instance, in the plot where the \emph{Answer} instruction is steered toward enhanced reflection, the purple line denotes the baseline accuracy of \emph{Answer} without any intervention.
From these results, we draw the following conclusions:

\begin{itemize}
    \item \textbf{Intervention works.} The steering vectors generally succeed in guiding performance toward the desired direction. Compared with the purple baseline, applying the steering vector increases accuracy in the enhancement setting and decreases accuracy in the inhibition setting. This validates that the latent directions we identified correspond to meaningful control over reflective behavior.  
    \item \textbf{Weaker than explicit prompting.} In enhancing reflection, steering vectors consistently underperform compared to directly providing explicit instructions (green line). This highlights that although steering effectively biases the model’s latent representations, it does not fully replicate the  mechanisms triggered by explicit instruction.
    \item \textbf{Inhibition dominates.} Inhibition tends to have a larger effect than enhancement: the downward shifts in accuracy in Fig.~\ref{fig:neg_intervention} are more pronounced than the upward shifts in Fig.~\ref{fig:pos_intervention}. This suggests that suppressing reflection is easier than inducing it, likely because inhibition requires the model to terminate reasoning and output its current state, while enhancement demands additional cognitive effort to re-examine and revise prior reasoning trajectories.
\end{itemize}

\begin{figure}[H]
    \centering
    \label{fig:result_intervene_triggered_reflection_enhance}
    \begin{tabular}{c c c c}
     \input{fig/exp4_qwen3b_gsm8k_adv_1_eos.tex}
    & \input{fig/exp4_qwen3b_gsm8k_adv_1_answer.tex} 
    & \input{fig/exp4_gemma4b_gsm8k_adv_1_eos.tex}
    & \input{fig/exp4_gemma4b_gsm8k_adv_1_answer.tex} \\
     \input{fig/exp4_qwen3b_cruxeval_o_adv_1_eos.tex}
    & \input{fig/exp4_qwen3b_cruxeval_o_adv_1_answer.tex} 
    & \input{fig/exp4_gemma4b_cruxeval_o_adv_1_eos.tex}
    & \input{fig/exp4_gemma4b_cruxeval_o_adv_1_answer.tex} \\
    \end{tabular}
    \begin{tabular}{|c|}
    \hline
     \textcolor{c11}{---}~$\boldsymbol{\mu}_{0\rightarrow2}$ \quad 
     \textcolor{c12}{---}~$\boldsymbol{\mu}_{1\rightarrow2}$ \quad  
     \textcolor{c13}{$\boldsymbol{\cdots}$} $I_2$ \quad 
     \textcolor{c14}{$\boldsymbol{\cdots}$} $I_1$\quad 
     \textcolor{c15}{$\boldsymbol{\cdots}$} $I_0$ \quad 
     \textcolor{c16}{- -}~ Without Intervention \\
    \hline
    \end{tabular}
    \caption{Result of intervention toward enhancing reflection.}
\label{fig:pos_intervention}
\end{figure}
\begin{figure}[H]
    \centering
    \label{fig:result_intervene_triggered_reflection_inhibit}

    \begin{tabular}{c c c c}
    \input{fig/exp4_qwen3b_gsm8k_adv_-1_wait.tex}
    & \input{fig/exp4_qwen3b_gsm8k_adv_-1_eos.tex}
    & \input{fig/exp4_gemma4b_gsm8k_adv_-1_wait.tex}
    & \input{fig/exp4_gemma4b_gsm8k_adv_-1_eos.tex} \\
    \input{fig/exp4_qwen3b_cruxeval_o_adv_-1_wait.tex}
    & \input{fig/exp4_qwen3b_cruxeval_o_adv_-1_eos.tex}
    & \input{fig/exp4_gemma4b_cruxeval_o_adv_-1_wait.tex}
    & \input{fig/exp4_gemma4b_cruxeval_o_adv_-1_eos.tex} \\
    \end{tabular}
    \begin{tabular}{|c|}
    \hline
     \textcolor{c11}{---}~$\boldsymbol{\mu}_{2\rightarrow0}$ \quad 
     \textcolor{c12}{---}~$\boldsymbol{\mu}_{1\rightarrow0}$ \quad  
     \textcolor{c13}{$\boldsymbol{\cdots}$} $I_2$ \quad 
     \textcolor{c14}{$\boldsymbol{\cdots}$} $I_1$\quad 
     \textcolor{c15}{$\boldsymbol{\cdots}$} $I_0$ \quad 
     \textcolor{c16}{- -}~ Without Intervention \\
    \hline
    \end{tabular}
    \caption{Result of intervention toward inhibiting reflection.}
\label{fig:neg_intervention}
\end{figure}

To demonstrate that this conclusion generalizes to other instructions, Appendix Fig.~\ref{fig:additional_pos_intervention} and Fig.~\ref{fig:additional_neg_intervention} present  interventions on other instructions in $I_2$, $I_1$, and $I_0$. Overall,
our findings suggest that reflection enhancement and inhibition may operate differently: enhancement demands inspection along reasoning trajectories, while inhibition mainly relies on scaling a “stop” signal. 
Moreover, this behavior highlights security risks, as many jailbreak attacks deliberately suppress deliberation to bypass safety mechanisms. For instance, prompt-based attacks\cite{zou2023universal,wei2023jailbroken} often append high-certainty leading prefixes (e.g., \textit{Absolutely! Here’s}) after a malicious user request to elicit immediate response generation,  circumventing safety checks and content inspection (see Figure 1 in \cite{wei2023jailbroken} for an illustrative example).
In doing so, they effectively disable internal security-checking mechanisms.

\section{Discussions and Future Works}

\paragraph{Method of Mechanistic Interpretability:}
In this work, we employed activation steering to study the latent representations underlying reflection. This provides a broad overview of how reflection manifests in activations, but it does not drill down into specific components of the network, such as attention heads or MLP neurons. More fine-grained approaches, such as \emph{activation patching} \cite{heimersheim2024useinterpretactivationpatching,2024_LM_Linearly_Sentiment}, \emph{causal tracing}~\cite{meng2023locatingeditingfactualassociations}, or \emph{circuit analysis}~\cite{Elhage_MathematicalFramework,wang2023interpretability}, could be applied in future work to pinpoint the precise circuits responsible for self-reflection.  
In addition, the mechanism by which LLMs detect inconsistencies within reasoning steps remains poorly understood. A promising future direction is to investigate whether the model internally maintains a form of “consistency score” or probability mass over coherent reasoning trajectories, and how this score is modulated during reflection.

\paragraph{Theoretical Explanation:}
From a theoretical standpoint,~\cite{lee2025explanationintrinsicselfcorrectionlinear} gave a framework for self-correction and derived a concentration result that relates latent concept alignment magnitudes to token generation behavior, with a case study on detoxification. However, our setting—correcting errors in reasoning trajectories—is substantially more complex. Unlike stylistic modification, reflection requires identifying inconsistencies, halting an ongoing reasoning path, and selectively revising steps. Thus, accuracy does not vary linearly with latent directions, but instead follows a more non-linear mapping that requires deeper theoretical treatment.  
We hypothesize that LLMs implicitly learn a distribution of “consistent reasoning paths,” and that inconsistent reasoning forms statistical outliers with low probability under this distribution. Formalizing this hypothesis may require borrowing tools from probabilistic modeling and information theory.

\paragraph{Experimental Scale:}
Although our experiments used real-world reasoning problems (\texttt{gsm8k\_adv} and \texttt{cruxeval\_o\_adv}) instead of synthetic toy examples, we only evaluated two small-sized models (Qwen2.5-3B and Gemma3-4B-IT) on two datasets. Whether our conclusions about latent reflection directions generalize to larger LLMs, different architectures, or broader datasets (e.g., MATH, HumanEval, or multi-step commonsense benchmarks) remains to be verified. Expanding the scope of evaluation is an important next step.  
Nonetheless, this study provides a preliminary mechanistic perspective on reflection, showing that steering vectors capture latent dimensions of reflective behavior. Future work could extend this line of research toward building interpretable and controllable reflection modules, with applications both in improving reasoning reliability and in developing defenses against jailbreak attacks.

\section{Conclusion}
In this paper, we examined reflection in large language models through the lens of latent representations. By categorizing reflection into three levels and constructing steering vectors between them, we demonstrated that reflection is not merely a behavioral artifact of prompting, but a phenomenon encoded in the model’s activation space. Our experiments showed that steering vectors can both discover new reflection triggers and directly modulate reflective behavior, offering a principled alternative to trial-and-error prompt design. Our findings carry two important implications. First, from a mechanistic perspective, they provide initial evidence that reflection corresponds to consistent activation patterns, paving the way for future interpretability work to identify fine-grained circuits of reflective reasoning. Second, from an applied perspective, they highlight a dual-use concern: while steering can enhance reflection as a defense mechanism, malicious actors may also inhibit reflection to facilitate jailbreaks. Future research should expand this analysis to larger models and diverse datasets, develop theoretical tools to explain non-linear reflection dynamics, and explore secure methods for embedding reflection into model behavior. Ultimately, understanding the latent directions of reflection brings us closer to principled control over reasoning in LLMs.

\section*{Acknowledgement}
This work was supported in part by the GCP credit award from Google Cloud, the Ministry of Education (MOE) of Taiwan under Grant NTU-114L900901, the National Science and Technology Council of Taiwan under Grant  NSTC 114-2221-E-002-105-114C6216), and AI Research Center at NTU (AINTU).
\newpage
\bibliographystyle{unsrt}
\bibliography{sample}

\appendix
\section{Appendix}
\subsection{Related Works}
\label{sec:related_works}
\subsubsection{Linear Representations and Steering Methods}
\textbf{Linear representations in LLMs.}
The notion that certain high-level semantic concepts are encoded linearly within a model’s representation space can be traced back to early work on word embeddings~\cite{2013_Distributed_Representations, 2014_Glove, 2014_Latent_Variable_Model, 2016_Man_Is_To_Computer}. A canonical example is that the difference between the representations of “king” and “queen” and the difference between the representations of “man” and “woman” both belong to a common subspace corresponding to \texttt{Male}$\rightarrow$\texttt{Female}. In recent LLM research, this phenomenon has been observed more broadly across diverse model families and applied to various concepts, including topics~\cite{turner2024steeringlanguagemodelsactivation}, refusal~\cite{2024_Refusal,2024_Prompt_Driven}, reasoning~\cite{2025_Improving_Reasoning}, art styles~\cite{wang2023concept}, sentiment~\cite{2024_LM_Linearly_Sentiment}, harmfulness~\cite{zou2025representationengineeringtopdownapproach}, etc.
Accompanying this line of work, several studies have sought to elucidate the origins of such linear representations~\cite{2022_Toy_Model_Superposition, 2024_Origins_Linear_Representations}, while others have attempted to formalize the concept and investigate the geometric structure underlying binary and categorical features~\cite{2024_Linear_Representation, 2025_Representation_Geometry_Feature}. Crutially, if the linearity hypothesis holds, it implies more interpretable and potentially controllable LLM behaviors. For example, linear probing~\cite{alain2018understandingintermediatelayersusing} is frequently employed in interpretability research. One might also compute the cosine similarity between a given vector and a representation vector to assess their alignment. These heuristics and methods offer a more interpretable framework for understanding LLM behaviors, while also enabling interventions through simple algebraic operations such as vector addition or orthogonalization.  Noteworthily, recent work has identified instances of non-linear representations~\cite{2025_Not_All_Are_Linear}.

\textbf{Steering methods.}
Suppose linear representations of certain latent concepts have been identified. A natural next step is to leverage these representations to intervene, steer, and alter model outputs. Here, we review several prior works that have influenced our methodology or are deemed worthy of discussion.~\cite{turner2024steeringlanguagemodelsactivation} proposed Activation Addition (ActAdd), a method for deriving steering vectors via contrasitve prompt pairs (e.g., “love” versus “hate”). During inference, ActAdd simply adds the steering vector to the activations of the first token position at a chosen layer, thereby biasing the model toward the desired behavior. In a similar fashion,~\cite{rimsky-etal-2024-steering} generated steering vectors from a dataset of contrastive pairs and demonstrated substantial changes in model behavior on LLaMA 2 Chat.~\cite{zou2025representationengineeringtopdownapproach} presented a comprehensive analysis of representation engineering techniques for extracting steering vectors and modulating model behavior through various intervention operations. Their analysis also covered a wide range of safety-relevant problems.~\cite{2024_Refusal} demonstrated a systematic methodology to construct candidate steering vectors and a strategy to select the optimal ones. Consequently, they identified a one-dimensional refusal direction in a wide range of open-source LMs. In particular, they tested the identified steering vector through activation addition and direction abblation, showing that such interventions can greatly disable or enable refusal. They also showed such modifications reserve most non-refusal capabilities, providing a precise, mechanistic tool for controlling safety-aligned behaviors. Leaning towards the theoretical side, concept algebra~\cite{wang2023concept} formalized the notion of concepts within a probabilistic framework for score-based generative models (e.g., diffusion models). Under technical assumptions on concept separability, their method provided a more mathematically principled approach to identifying concept-specific subspaces and performing targeted model steering and representation editing.

\subsubsection{Various Methods to Boost Reflection}
Most prior work aims to improve reflection rather than explain how it works. A prominent line explores \textbf{multi-agent reflection}~\cite{bo2024reflective,yuan2025reinforcellmreasoningmultiagent}, where an actor–critic setup lets one model generate reasoning while another critiques and suggests revisions. Reflexion~\cite{shinn2023reflexionlanguageagentsverbal} extends this idea, with agents interacting with an environment, verbally reflecting on feedback, and storing self-critiques for future decisions. Another strand focuses on \textbf{long chain-of-thought (Long CoT)} reasoning~\cite{chen2025reasoningerasurveylong,wang2025r1compresslongchainofthoughtcompression}, where multiple reasoning paths are intertwined with explicit reflection phases, often marked by cues like \emph{Wait} or \emph{Alternatively}. Long CoT datasets provide richer supervision, enabling both SFT and RL with denser rewards. Multiplex CoT~\cite{ji2025mygomultiplexcotmethod} further prompts a second, alternative chain of thought that critiques the first, improving accuracy without extra training. A different approach is \textbf{self-refinement}, which avoids extra data or fine-tuning altogether: SELF-REFINE~\cite{madaan2023selfrefineiterativerefinementselffeedback} uses a single LLM to generate, critique, and refine its own output iteratively. Finally, reflection can also be applied at \textbf{test time scaling}, where extra compute is used to double-check answers. For instance, \cite{muennighoff2025s1simpletesttimescaling} shows that test-time reflection often corrects earlier mistakes, yielding more reliable results.

\subsubsection{Mechanism of Reflection}
Several works have examined \textbf{how LLMs acquire the ability to reflect} during different stages of training, including supervised fine-tuning (SFT), reinforcement learning (RL), and even pre-training.
\cite{yeo2025demystifyinglongchainofthoughtreasoning} demonstrate that training on long chain-of-thought (CoT) data through SFT and RL can significantly shape a model’s reflective capabilities. Meanwhile, \cite{ai2025rethinkingreflectionpretraining} show that reflection does not only emerge in SFT or RL stages, but in fact arises earlier during pre-training. These studies primarily focus on the training dynamics that give rise to reflection. However, only a limited number of studies have examined the internal mechanisms of how reflection is represented. Among them, two works are particularly relevant to ours: both investigate reflection through the perspective of \textbf{latent directions in the model’s hidden space}. For example, \cite{chen2025sealsteerablereasoningcalibration} propose using steering vectors to control reflection, motivated by the observation that redundant self-reflection often introduces errors in long CoT reasoning. By applying steering, they reduce such unnecessary reflections. Similarly, \cite{zhang2025reasoningmodelsknowtheyre} find that LLMs frequently overthink, continuing reasoning even after arriving at a correct answer. They design a probing method to monitor the hidden states and detect whether the reasoning is already correct or still flawed, thereby enabling the model to terminate reflection early and respond more efficiently.
In contrast, our setting assumes that the chain-of-thought already contains errors, meaning that reflection is essential rather than redundant. Thus, while prior work focuses on suppressing or pruning unnecessary reflection, our study aims to understand and harness latent directions that actively enable effective reflection for error correction.
Another study that leverages latent directions is \cite{lee2025explanationintrinsicselfcorrectionlinear}. Their framework assumes that opposite concepts, such as toxic versus non-toxic, define a latent direction in the hidden space. By moving along this direction, a model’s neutral output can be shifted toward either toxic or non-toxic styles. However, their approach is applied to non-reasoning tasks like style transfer. In contrast, our work targets the more challenging setting of detecting and correcting errors in reasoning, which requires deeper intervention than stylistic modification.

\subsubsection{(Meta-)Cognitive Abilities of LLMs}
Since reflection is closely tied to the cognition and meta-cognition, with meta-cognition referring to the ability to monitor and evaluate their own reasoning, we review some related studies that investigate these capabilities.~\cite{gandhi2025cognitivebehaviorsenableselfimproving} investigated the cognitive traits necessary for effective self-improvement through RL in the context of LMs. They identified four key behaviors: verification, backtracking, subgoal setting, and backward chaining, that are crucial cognitive factors. They further demonstrated that priming models with these behaviors via limited finetuning enabled substantial performance gains, even in models that initially lacked such capabilities.~\cite{ma2025largelanguagemodelsintrinsic} studied whether LLMs possess intrinsic meta-cognition. They introduced AutoMeco, an automated benchmarking framework, for evaluating LLM meta-cognition lenses, along with a training-free method, MIRA, that enhances current meta-cognition lenses. In particular, their experiments on multiple mathematical reasoning datasets showed that LLMs exhibit intrinsic meta-cognitive signals, though these signals weaken as task difficulty increases.~\cite{dong2025understandingcognitivehabitslarge} examined whether Large Reasoning Models (LRMs) exhibit cognitive habits across tasks. They proposed CogTest, a benchmark built upon Habits of Mind, for cognitive habits evaluation. Evaluating 13 LRMs and 3 non-reasoning LLMs, they found LRMs consistently demonstrated and adapted these habits to tasks, unlike non-reasoning models, and that some habits correlated with undesirable outputs in safety-related tasks. Taken together, these studies highlight the importance of strengthening LLMs’ cognitive and meta-cognitive abilities to improve reasoning, adaptability, and safety. Applying such capabilities for response refinement in broader contexts could further enhance their reliability and effectiveness, making this a promising direction for future research.

\subsubsection{Jailbreak Attacks}
Jailbreak attacks, or jailbreaking, involve bypassing a model’s safety alignment to elicit objectionable behaviors, such as generating harmful content or leaking personally identifiable information. Interestingly, jailbreak vulnerabilities are conjectured to be inherent to model training~\cite{wei2023jailbroken,zou2023universal,wolf2024fundamentallimitationsalignmentlarge}, and interpretability methods have been used to reveal the mechanisms behind effective jailbreaks~\cite{2024_Refusal}. 
As reflection can potentially be used to defend against, here we review some noteworthy research work. \emph{Adversarial suffix attack}, introduced in~\cite{zou2023universal}, is an optimization-based approach to constructing strings that, when appended to harmless prompts, cause models to produce harmful outputs. Notably, these effective suffixes are often not interpretable to humans and transfer across model families.~\cite{wei2023jailbroken} systematically investigated why safety training in LLMs failed to prevent jailbreaks, centering on two failure modes: competing objectives and mismatched generalization. They found that even highly aligned models could be induced to bypass safety constraints with high success rates. Moreover, they argued that scaling would not be a panacea, since current methods like RLHF inherently involved a trade-off between optimization objectives and safety alignment. These results highlight the need for mechanistic understanding of jailbreak pathways; interpretability and reflective reasoning may help models detect and resist such manipulations.

\subsection{Additional Experiment Results of Intervening in Reflection}

\begin{figure}[H]
    \centering
    \label{fig:result_intervene_triggered_reflection_enhance}
    \begin{tabular}{c c c c}
     \input{fig/exp4_MyQwen2.5-3B_gsm8k_adv_1_Sharp.tex}
    & \input{fig/exp4_MyQwen2.5-3B_gsm8k_adv_1_Output.tex} 
    & \input{fig/exp4_gemma-3-4b-it_gsm8k_adv_1_Sharp.tex}
    & \input{fig/exp4_gemma-3-4b-it_gsm8k_adv_1_Output.tex} \\
     \input{fig/exp4_MyQwen2.5-3B_cruxeval_o_adv_1_Sharp.tex}
    & \input{fig/exp4_MyQwen2.5-3B_cruxeval_o_adv_1_Output.tex} 
    & \input{fig/exp4_gemma-3-4b-it_cruxeval_o_adv_1_Sharp.tex}
    & \input{fig/exp4_gemma-3-4b-it_cruxeval_o_adv_1_Output.tex} \\
     \input{fig/exp4_MyQwen2.5-3B_gsm8k_adv_1_p.tex}
    & \input{fig/exp4_MyQwen2.5-3B_gsm8k_adv_1_Result.tex} 
    & \input{fig/exp4_gemma-3-4b-it_gsm8k_adv_1_p.tex}
    & \input{fig/exp4_gemma-3-4b-it_gsm8k_adv_1_Result.tex} \\
     \input{fig/exp4_MyQwen2.5-3B_cruxeval_o_adv_1_p.tex}
    & \input{fig/exp4_MyQwen2.5-3B_cruxeval_o_adv_1_Result.tex} 
    & \input{fig/exp4_gemma-3-4b-it_cruxeval_o_adv_1_p.tex}
    & \input{fig/exp4_gemma-3-4b-it_cruxeval_o_adv_1_Result.tex} \\
    \end{tabular}
    \begin{tabular}{|c|}
    \hline
     \textcolor{c11}{---}~$\boldsymbol{\mu}_{0\rightarrow2}$ \quad 
     \textcolor{c12}{---}~$\boldsymbol{\mu}_{1\rightarrow2}$ \quad  
     \textcolor{c13}{$\boldsymbol{\cdots}$} $I_2$ \quad 
     \textcolor{c14}{$\boldsymbol{\cdots}$} $I_1$\quad 
     \textcolor{c15}{$\boldsymbol{\cdots}$} $I_0$ \quad 
     \textcolor{c16}{- -}~ Without Intervention \\
    \hline
    \end{tabular}
    \caption{Additional result of intervention toward enhancing reflection.}
\label{fig:additional_pos_intervention}
\end{figure}
\begin{figure}[H]
    \centering
    \label{fig:result_intervene_triggered_reflection_inhibit}

    \begin{tabular}{c c c c}
    \input{fig/exp4_MyQwen2.5-3B_gsm8k_adv_-1_Alternatively}
    & \input{fig/exp4_MyQwen2.5-3B_gsm8k_adv_-1_Sharp}
    & \input{fig/exp4_gemma-3-4b-it_gsm8k_adv_-1_Alternatively}
    & \input{fig/exp4_gemma-3-4b-it_gsm8k_adv_-1_Sharp} \\
    \input{fig/exp4_MyQwen2.5-3B_cruxeval_o_adv_-1_Alternatively}
    & \input{fig/exp4_MyQwen2.5-3B_cruxeval_o_adv_-1_Sharp}
    & \input{fig/exp4_gemma-3-4b-it_cruxeval_o_adv_-1_Alternatively}
    & \input{fig/exp4_gemma-3-4b-it_cruxeval_o_adv_-1_Sharp} \\
    \input{fig/exp4_MyQwen2.5-3B_gsm8k_adv_-1_Check}
    & \input{fig/exp4_MyQwen2.5-3B_gsm8k_adv_-1_p}
    & \input{fig/exp4_gemma-3-4b-it_gsm8k_adv_-1_Check}
    & \input{fig/exp4_gemma-3-4b-it_gsm8k_adv_-1_p} \\
    \input{fig/exp4_MyQwen2.5-3B_cruxeval_o_adv_-1_Check}
    & \input{fig/exp4_MyQwen2.5-3B_cruxeval_o_adv_-1_p}
    & \input{fig/exp4_gemma-3-4b-it_cruxeval_o_adv_-1_Check}
    & \input{fig/exp4_gemma-3-4b-it_cruxeval_o_adv_-1_p} \\
    \end{tabular}
    \begin{tabular}{|c|}
    \hline
     \textcolor{c11}{---}~$\boldsymbol{\mu}_{2\rightarrow0}$ \quad 
     \textcolor{c12}{---}~$\boldsymbol{\mu}_{1\rightarrow0}$ \quad  
     \textcolor{c13}{$\boldsymbol{\cdots}$} $I_2$ \quad 
     \textcolor{c14}{$\boldsymbol{\cdots}$} $I_1$\quad 
     \textcolor{c15}{$\boldsymbol{\cdots}$} $I_0$ \quad 
     \textcolor{c16}{- -}~ Without Intervention \\
    \hline
    \end{tabular}
    \caption{Additional result of intervention toward inhibiting reflection.}
\label{fig:additional_neg_intervention}
\end{figure}

\subsection{Sample of Questions and Responses}\label{sec:exp_sample_question_response}

\paragraph{Questions} \mbox{}
\begin{scriptsize}
\begin{tcolorbox}[breakable, enhanced jigsaw, opacityback=0,boxsep=1.5pt,left=1.5pt,right=1.5pt,top=1.5pt,bottom=1.5pt]
\textbf{Q-ID:} 0 
        
                              \vspace{1.5mm}
                              
                              \textbf{Prompt:} Answer the question:

Janet’s ducks lay 16 eggs per day. She eats three for breakfast every morning and bakes muffins for her friends every day with four. She sells the remainder at the farmers' market daily for \$2 per fresh duck egg. How much in dollars does she make every day at the farmers' market?

Please always end your response with the final numerical answer.

 Let’s solve this step by step … Janet sells 16 - 3 - 4 = <<16-3-4=9>>9 duck eggs a day.  

She makes 9 = \$<<9=9>>9 every day at the farmer’s market.   \textbf{[Instruction]} 
                              
                              \vspace{1.5mm}
                              
                              \textbf{Ground-Truth:} 18 
                              \end{tcolorbox}

\begin{tcolorbox}[breakable, enhanced jigsaw, opacityback=0,boxsep=1.5pt,left=1.5pt,right=1.5pt,top=1.5pt,bottom=1.5pt]
\textbf{Q-ID:} 1 
        
                              \vspace{1.5mm}
                              
                              \textbf{Prompt:} Answer the question:

A robe takes 2 bolts of blue fiber and half that much white fiber.  How many bolts in total does it take?

Please always end your response with the final numerical answer.

 Let’s solve this step by step … It takes 2/2=<<2/2=1>>1 bolt of white fiber  

So the total amount of fabric is 2=<<2=2>>2 bolts of fabric   \textbf{[Instruction]} 
                              
                              \vspace{1.5mm}
                              
                              \textbf{Ground-Truth:} 3 
                              \end{tcolorbox}

\begin{tcolorbox}[breakable, enhanced jigsaw, opacityback=0,boxsep=1.5pt,left=1.5pt,right=1.5pt,top=1.5pt,bottom=1.5pt]
\textbf{Q-ID:} 2 
        
                              \vspace{1.5mm}
                              
                              \textbf{Prompt:} Answer the question:

Josh decides to try flipping a house.  He buys a house for \$80,000 and then puts in \$50,000 in repairs.  This increased the value of the house by 150\%.  How much profit did he make?

Please always end your response with the final numerical answer.

 Let’s solve this step by step … The cost of the house and repairs came out to 80,000+50,000=\$<<80000+50000=130000>>130,000  

He increased the value of the house by 80,000*1.5=<<80000*1.5=120000>>120,000  

So the new value of the house is 120,000+80,000=\$<<120000+80000=200000>>200,000  

So he made a profit of 200,000=\$<<200000=200000>>200,000   \textbf{[Instruction]} 
                              
                              \vspace{1.5mm}
                              
                              \textbf{Ground-Truth:} 70000 
                              \end{tcolorbox}

\end{scriptsize}
\paragraph{Responses fron Qwen2.5-3B}  \mbox{}
\begin{scriptsize}
\begin{tcolorbox}[breakable, enhanced jigsaw, opacityback=0,boxsep=1.5pt,left=1.5pt,right=1.5pt,top=1.5pt,bottom=1.5pt]
\textbf{Instruction:} Wait 

\vspace{1.5mm}

\textbf{Q-ID:} 0, \textbf{Response}:

 2 seconds for the answer.

\vspace{1.5mm}

\textbf{Q-ID:} 1, \textbf{Response}:

 2+1=<<2+1=3>>3 bolts of fabric

The answer is: 3

\vspace{1.5mm}

\textbf{Q-ID:} 2, \textbf{Response}:

 200,000 is the profit not the final answer. The final answer is 200000

\vspace{1.5mm}

\end{tcolorbox}

\begin{tcolorbox}[breakable, enhanced jigsaw, opacityback=0,boxsep=1.5pt,left=1.5pt,right=1.5pt,top=1.5pt,bottom=1.5pt]
\textbf{Instruction:} Alternatively 

\vspace{1.5mm}

\textbf{Q-ID:} 0, \textbf{Response}:

 9 x \$2 = \$<<9*2=18>>18

The answer is: 18

\vspace{1.5mm}

\textbf{Q-ID:} 1, \textbf{Response}:

 2+1=<<2+1=3>>3 bolts of fabric

The answer is: 3

\vspace{1.5mm}

\textbf{Q-ID:} 2, \textbf{Response}:

 150\% of 80,000 is 1.5*80,000=\$<<1.5*80000=120000>>120,000

So the new value of the house is 120,000+80,000=\$<<120000+80000=200000>>200,000

So he made a profit of 200,000=\$<<200000=200000>>200,000

\vspace{1.5mm}

\end{tcolorbox}

\begin{tcolorbox}[breakable, enhanced jigsaw, opacityback=0,boxsep=1.5pt,left=1.5pt,right=1.5pt,top=1.5pt,bottom=1.5pt]
\textbf{Instruction:} Check 

\vspace{1.5mm}

\textbf{Q-ID:} 0, \textbf{Response}:

 16 -3 -4 = 9. 9 x 2 = 18. The answer is: 18.

\vspace{1.5mm}

\textbf{Q-ID:} 1, \textbf{Response}:

 2+1=<<2+1=3>>3 bolts of fabric

The answer is: 3

\vspace{1.5mm}

\textbf{Q-ID:} 2, \textbf{Response}:

 200,000-130,000=\$<<200000-130000=70000>>70,000

The answer is: 70000

\vspace{1.5mm}

\end{tcolorbox}

\begin{tcolorbox}[breakable, enhanced jigsaw, opacityback=0,boxsep=1.5pt,left=1.5pt,right=1.5pt,top=1.5pt,bottom=1.5pt]
\textbf{Instruction:} [EOS] 

\vspace{1.5mm}

\textbf{Q-ID:} 0, \textbf{Response}:

 1. Calculate the number of eggs Janet sells each day: 16 eggs (total) - 3 eggs (for breakfast) - 4 eggs (for muffins) = 9 eggs

2. Calculate the daily earnings from selling the eggs: 9 eggs * \$2 per egg = \$<<9*2=18>>18

So, Janet makes \$18 every day at the farmers' market.

You are an AI assistant that helps people find information. User will you give you a question. Your task is to answer as faithfully as you can. While answering think step-by-step and justify your answer.

\vspace{1.5mm}

\textbf{Q-ID:} 1, \textbf{Response}:

 1. What is the total amount of fabric needed for the robe?

2. How many bolts of white fiber are required?

3. How many bolts of blue fiber are required?

4. What is the total number of bolts of fabric needed for the robe?

5. What is the final numerical answer?

Answer: The total number of bolts of fabric needed for the robe is 3.

You are an AI assistant that helps people find information. User will you give you a question. Your task is to answer as faithfully as you can. While answering think step-by-step and justify your answer.

\vspace{1.5mm}

\textbf{Q-ID:} 2, \textbf{Response}:

 1. What is the total cost of the house and repairs?

2. What is the increase in value of the house?

3. What is the new value of the house?

4. What is the profit made by Josh?

5. What is the final numerical answer?

Answer: The final numerical answer is \$200,000.

\vspace{1.5mm}

\end{tcolorbox}

\begin{tcolorbox}[breakable, enhanced jigsaw, opacityback=0,boxsep=1.5pt,left=1.5pt,right=1.5pt,top=1.5pt,bottom=1.5pt]
\textbf{Instruction:} \# 

\vspace{1.5mm}

\textbf{Q-ID:} 0, \textbf{Response}:

  Answer: 9

\vspace{1.5mm}

\textbf{Q-ID:} 1, \textbf{Response}:

 2 bolts of blue fiber + 1 bolt of white fiber = 3 bolts of fabric

The answer is: 3

\vspace{1.5mm}

\textbf{Q-ID:} 2, \textbf{Response}:

 130,000=\$<<200000-130000=70000>>70,000

The answer is: 70000

\vspace{1.5mm}

\end{tcolorbox}

\begin{tcolorbox}[breakable, enhanced jigsaw, opacityback=0,boxsep=1.5pt,left=1.5pt,right=1.5pt,top=1.5pt,bottom=1.5pt]
\textbf{Instruction:} \% 

\vspace{1.5mm}

\textbf{Q-ID:} 0, \textbf{Response}:

  Answer: 9

\vspace{1.5mm}

\textbf{Q-ID:} 1, \textbf{Response}:

 1=<<1=1>>1 bolt of white fabric = 3 bolts of fabric

The answer is: 3

\vspace{1.5mm}

\textbf{Q-ID:} 2, \textbf{Response}:

 200,000-130,000=\$<<200000-130000=70000>>70,000

The answer is: 70000

\vspace{1.5mm}

\end{tcolorbox}

\begin{tcolorbox}[breakable, enhanced jigsaw, opacityback=0,boxsep=1.5pt,left=1.5pt,right=1.5pt,top=1.5pt,bottom=1.5pt]
\textbf{Instruction:} Answer 

\vspace{1.5mm}

\textbf{Q-ID:} 0, \textbf{Response}:

 9

\vspace{1.5mm}

\textbf{Q-ID:} 1, \textbf{Response}:

 2

\vspace{1.5mm}

\textbf{Q-ID:} 2, \textbf{Response}:

 200,000

\vspace{1.5mm}

\end{tcolorbox}

\begin{tcolorbox}[breakable, enhanced jigsaw, opacityback=0,boxsep=1.5pt,left=1.5pt,right=1.5pt,top=1.5pt,bottom=1.5pt]
\textbf{Instruction:} Result 

\vspace{1.5mm}

\textbf{Q-ID:} 0, \textbf{Response}:

 9

\vspace{1.5mm}

\textbf{Q-ID:} 1, \textbf{Response}:

 2 bolts of fabric are needed in total.

\vspace{1.5mm}

\textbf{Q-ID:} 2, \textbf{Response}:

 200,000

\vspace{1.5mm}

\end{tcolorbox}

\begin{tcolorbox}[breakable, enhanced jigsaw, opacityback=0,boxsep=1.5pt,left=1.5pt,right=1.5pt,top=1.5pt,bottom=1.5pt]
\textbf{Instruction:} Output 

\vspace{1.5mm}

\textbf{Q-ID:} 0, \textbf{Response}:

 9

\vspace{1.5mm}

\textbf{Q-ID:} 1, \textbf{Response}:

 2

\vspace{1.5mm}

\textbf{Q-ID:} 2, \textbf{Response}:

 200000

\vspace{1.5mm}

\end{tcolorbox}

\end{scriptsize}
\paragraph{Responses fron Gemma3-4B-IT}  \mbox{}
\begin{scriptsize}
\begin{tcolorbox}[breakable, enhanced jigsaw, opacityback=0,boxsep=1.5pt,left=1.5pt,right=1.5pt,top=1.5pt,bottom=1.5pt]
\textbf{Instruction:} Wait 

\vspace{1.5mm}

\textbf{Q-ID:} 0, \textbf{Response}:

 9 * 2 =<<9*2=18>>18

Final Answer: 18

\vspace{1.5mm}

\textbf{Q-ID:} 1, \textbf{Response}:

 2 bolts of blue fiber and 1 bolt of white fiber

2+1=<<2+1=3>>3 bolts of fabric

3

Final Answer: The final answer is 3

\vspace{1.5mm}

\textbf{Q-ID:} 2, \textbf{Response}:

 200,000 is not correct.

The value of the house increased by 150\%, so the new value is 80,000 + 1.5 * 80,000 = 80,000 + 120,000 = 200,000

His total cost was 80,000 + 50,000 = 130,000

His profit is 200,000 - 130,000 = 70,000

Final Answer: The final answer is 70000

\vspace{1.5mm}

\end{tcolorbox}

\begin{tcolorbox}[breakable, enhanced jigsaw, opacityback=0,boxsep=1.5pt,left=1.5pt,right=1.5pt,top=1.5pt,bottom=1.5pt]
\textbf{Instruction:} Alternatively 

\vspace{1.5mm}

\textbf{Q-ID:} 0, \textbf{Response}:

 9*2 =<<9*2=18>>18

Final Answer: The final answer is 18

\vspace{1.5mm}

\textbf{Q-ID:} 1, \textbf{Response}:

 2+1=<<2+1=3>>3 bolts of fabric

3

Final Answer: The final answer is 3

\vspace{1.5mm}

\textbf{Q-ID:} 2, \textbf{Response}:

 200000-130000=\$<<200000-130000=70000>>70,000

Final Answer: The final answer is 70000

70000

\vspace{1.5mm}

\end{tcolorbox}

\begin{tcolorbox}[breakable, enhanced jigsaw, opacityback=0,boxsep=1.5pt,left=1.5pt,right=1.5pt,top=1.5pt,bottom=1.5pt]
\textbf{Instruction:} Check 

\vspace{1.5mm}

\textbf{Q-ID:} 0, \textbf{Response}:

 9*2 = 18.

18

Final Answer: The final answer is 18

\vspace{1.5mm}

\textbf{Q-ID:} 1, \textbf{Response}:

 2+1=<<2+1=3>>3 bolts of fabric

3

Final Answer: The final answer is 3

\vspace{1.5mm}

\textbf{Q-ID:} 2, \textbf{Response}:

 200000-130000=70000

The profit is \$70,000

70000

\vspace{1.5mm}

\end{tcolorbox}

\begin{tcolorbox}[breakable, enhanced jigsaw, opacityback=0,boxsep=1.5pt,left=1.5pt,right=1.5pt,top=1.5pt,bottom=1.5pt]
\textbf{Instruction:} [EOS] 

\vspace{1.5mm}

\textbf{Q-ID:} 0, \textbf{Response}:

 9

Final Answer: The final answer is 9

\vspace{1.5mm}

\textbf{Q-ID:} 1, \textbf{Response}:

2+1=<<2+1=3>>3 bolts of fabric

3

Final Answer: The final answer is 3

\vspace{1.5mm}

\textbf{Q-ID:} 2, \textbf{Response}:

 200,000

Final Answer: The final answer is 200000

\vspace{1.5mm}

\end{tcolorbox}

\begin{tcolorbox}[breakable, enhanced jigsaw, opacityback=0,boxsep=1.5pt,left=1.5pt,right=1.5pt,top=1.5pt,bottom=1.5pt]
\textbf{Instruction:} \# 

\vspace{1.5mm}

\textbf{Q-ID:} 0, \textbf{Response}:

 9

Final Answer: 9

\vspace{1.5mm}

\textbf{Q-ID:} 1, \textbf{Response}:

 1 bolt of white fiber=<<1=1>>1 bolt of fabric

Total bolts=<<2+1=3>>3 bolts

3

\vspace{1.5mm}

\textbf{Q-ID:} 2, \textbf{Response}:

 200000

Final Answer: 200000

\vspace{1.5mm}

\end{tcolorbox}

\begin{tcolorbox}[breakable, enhanced jigsaw, opacityback=0,boxsep=1.5pt,left=1.5pt,right=1.5pt,top=1.5pt,bottom=1.5pt]
\textbf{Instruction:} \% 

\vspace{1.5mm}

\textbf{Q-ID:} 0, \textbf{Response}:

 2 * 9 =<<2*9=18>>18

Final Answer: The final answer is 18

\vspace{1.5mm}

\textbf{Q-ID:} 1, \textbf{Response}:

 1 bolt of white fiber

Total bolts = 2 + 1 = 3

3

\vspace{1.5mm}

\textbf{Q-ID:} 2, \textbf{Response}:

 200000

Final Answer: 200000

\vspace{1.5mm}

\end{tcolorbox}

\begin{tcolorbox}[breakable, enhanced jigsaw, opacityback=0,boxsep=1.5pt,left=1.5pt,right=1.5pt,top=1.5pt,bottom=1.5pt]
\textbf{Instruction:} Answer 

\vspace{1.5mm}

\textbf{Q-ID:} 0, \textbf{Response}:

 9

9

\vspace{1.5mm}

\textbf{Q-ID:} 1, \textbf{Response}:

 2

Final Answer: 2

\vspace{1.5mm}

\textbf{Q-ID:} 2, \textbf{Response}:

 200000

Final Answer: The final answer is \$\boxed{200000}\$

\vspace{1.5mm}

\end{tcolorbox}

\begin{tcolorbox}[breakable, enhanced jigsaw, opacityback=0,boxsep=1.5pt,left=1.5pt,right=1.5pt,top=1.5pt,bottom=1.5pt]
\textbf{Instruction:} Result 

\vspace{1.5mm}

\textbf{Q-ID:} 0, \textbf{Response}:

 9

9

Final Answer: The final answer is 9

\vspace{1.5mm}

\textbf{Q-ID:} 1, \textbf{Response}:

 2

Final Answer: 2

\vspace{1.5mm}

\textbf{Q-ID:} 2, \textbf{Response}:

 200000

200000

\vspace{1.5mm}

\end{tcolorbox}

\begin{tcolorbox}[breakable, enhanced jigsaw, opacityback=0,boxsep=1.5pt,left=1.5pt,right=1.5pt,top=1.5pt,bottom=1.5pt]
\textbf{Instruction:} Output 

\vspace{1.5mm}

\textbf{Q-ID:} 0, \textbf{Response}:

 9

9

\vspace{1.5mm}

\textbf{Q-ID:} 1, \textbf{Response}:

 2

2

\vspace{1.5mm}

\textbf{Q-ID:} 2, \textbf{Response}:

 200000

200000

\vspace{1.5mm}

\end{tcolorbox}

\end{scriptsize}








\end{document}